\documentclass{article}

\usepackage{microtype}
\usepackage{graphicx}
\usepackage{subfigure}
\usepackage{booktabs} 
\usepackage{algorithm}

\usepackage{hyperref}

\usepackage[accepted]{icml2019}

\usepackage{multirow}
\usepackage{diagbox}
\usepackage{amsmath}
\usepackage{bm}

\usepackage{enumerate}
\usepackage{graphicx}

\newcommand{\best}[1]{{\textbf{\color{red}#1}}}
\newcommand{\second}[1]{{\underline{\color{blue}#1}}}

\usepackage{lipsum}
\newcommand\blfootnote[1]{%
  \begingroup
  \renewcommand\thefootnote{}\footnote{#1}%
  \addtocounter{footnote}{-1}%
  \endgroup
}

\icmltitlerunning{Unsupervised Deep Learning by Neighbourhood Discovery}

\begin{document}

\twocolumn[
\icmltitle{Unsupervised Deep Learning by Neighbourhood Discovery}



\icmlsetsymbol{equal}{*}

\begin{icmlauthorlist}
\icmlauthor{Jiabo Huang}{qmul}
\icmlauthor{Qi Dong}{qmul}
\icmlauthor{Shaogang Gong}{qmul}
\icmlauthor{Xiatian Zhu}{vsl}
\end{icmlauthorlist}

\icmlaffiliation{qmul}{Queen Mary University of London}
\icmlaffiliation{vsl}{Vision Semantics Limited}

\icmlcorrespondingauthor{Xiatian Zhu}{eddy.zhuxt@gmail.com}

\icmlkeywords{Machine Learning, ICML, Unsupervised Learning, Curriculum Learning}

\vskip 0.3in
]



\printAffiliationsAndNotice{}  

\begin{abstract}
Deep convolutional neural networks (CNNs)
have demonstrated remarkable success in computer vision
by {\em supervisedly} learning strong visual feature representations.
However, training CNNs relies heavily on the availability of exhaustive training
data annotations, limiting significantly their deployment and scalability
in many application scenarios.
In this work, we introduce a generic {\em unsupervised deep learning} 
approach to training deep models
without the need for any manual label supervision.
Specifically, we progressively discover sample anchored/centred
neighbourhoods 
to reason and learn the underlying class decision boundaries
iteratively and accumulatively.
Every single neighbourhood is specially formulated so that all the member samples
can share the same unseen class labels at high probability
for facilitating the extraction of 
class discriminative feature representations
during training.
Experiments on image classification show the performance 
advantages of the proposed method over the state-of-the-art
unsupervised learning models on six benchmarks
including both coarse-grained and fine-grained 
object image categorisation. 

\end{abstract}

\section{Introduction}
Deep neural networks, particularly convolutional neural networks (CNNs), have significantly
advanced the progress of computer vision problems \cite{goodfellow2016deep,lecun2015deep}.
However, such achievements are largely established upon 
{\em supervised learning} of network models
on a massive collection of exhaustively labelled training imagery data \cite{krizhevsky2012imagenet,dong2018single,dong2018imbalanced}.
This dramatically restricts their scalability and usability
to many practical applications with limited labelling budgets.
A natural solution is 
{\em unsupervised learning} of deep feature representations,
which has recently drawn increasing attention
\cite{cvpr2018wu_unsupervised,eccv2018Caron_deepclustering}.

In the literature, 
representative unsupervised deep learning methods 
include clustering \cite{eccv2018Caron_deepclustering,icml2016xie_unsupervised,icml2016yang_kmeans-friendly} and 
sample specificity analysis \cite{cvpr2018wu_unsupervised,icml2017bojanowski_nat}.
The objective of clustering is to identify a set of clusters
and each represents an underlying class concept.
This strategy has great potential with the best case reaching to 
the performance of supervised learning,   
but is error-prone due to the enormous combinatorial space
and complex class boundaries.
In contrast, sample specificity learning
avoids the cluster notion by treating every single sample
as an independent class. 
The hypothesis is that the model can reveal
the underlying class-to-class semantic similarity structure,
e.g. the manifold geometry.
Whilst collecting such instance labels requires no manual annotation cost,
the resulting supervision is ambiguous therefore weak to class discrimination.
Other contemporary self-supervised learning methods
\cite{iccv2015Gupta_context,eccv2016zhang_color,eccv2016noroozi_jigsaw,iccv2017noroozi_count,cvpr2017zhang_splitbrain}
share a similar limitation
due to the insufficient correlation between
the auxiliary supervision and the underlying class target.

\blfootnote{Code is available at \href{https://github.com/Raymond-sci/AND}{https://github.com/raymond-sci/AND}.}

In this work, we present a generic unsupervised deep learning 
method called {\em Anchor Neighbourhood Discovery} (AND).
The AND model combines the advantages of both
clustering and sample specificity learning whilst mitigating
their disadvantages in a principled formulation.
Specifically, with a {\em divide-and-conquer} principle,
the AND discovers class consistent neighbourhoods 
anchored to individual training samples ({\em divide})
and propagates the local inter-sample class relationships within such neighbourhoods
({\em conquer})
for more reliably extracting the latent discrimination information
during model training.
Neighbourhoods can be considered
as tiny sample anchored clusters with higher compactness 
and class consistency.
They are specially designed for minimising the clustering errors 
whilst retaining the exploration of inter-sample class information
that is entirely neglected in sample specificity learning.
To enhance the neighbourhood quality (class consistency),
we introduce a progressive discovery curriculum
for incrementally deriving more accurate neighbourhood supervision.

We make three {\bf contributions}:
{\bf (1)} We propose the idea of exploiting local neighbourhoods 
for unsupervised deep learning.
This strategy preserves the capability of clustering 
for class boundary inference whilst minimising the negative impact 
of class inconsistency typically encountered in clusters.
To our best knowledge, it is the first attempt at
exploring the concept of neighbourhood for
end-to-end deep learning of feature representations
without class label annotations.
{\bf (2)} We formulate an {\em Anchor Neighbourhood Discovery} (AND)
approach to progressive unsupervised deep learning.
The AND model not only generalises the idea of 
sample specificity learning,
but also additionally 
considers 
the originally missing sample-to-sample correlation
during model learning by a novel neighbourhood supervision design.  
{\bf (3)} 
We further introduce a curriculum learning algorithm
to gradually perform neighbourhood discovery for
maximising the class consistency
of neighbourhoods therefore enhancing the 
unsupervised learning capability.

Extensive experiments are conducted on
four coarse-grained
(CIFAR10 and CIFAR100~\cite{krizhevsky2009cifar}, 
SVHN~\cite{netzer2011svhn}, ImageNet~\cite{russakovsky2015imagenet})
and two fine-grained 
(CUB200-2011~\cite{Wah2011CUB200}
and Stanford Dogs~\cite{khosla2011dogs})
object image classification datasets.
The results show the advantages of our AND method 
over a wide variety of existing state-of-the-art unsupervised deep learning models.


\section{Related Work}
Existing unsupervised deep learning methods 
generally fall into four different categories:
{\bf(1)} Clustering analysis \cite{eccv2018Caron_deepclustering,icml2016xie_unsupervised,icml2016yang_kmeans-friendly},
{\bf(2)} Sample specificity learning 
\cite{cvpr2018wu_unsupervised,icml2017bojanowski_nat},
{\bf(3)} Self-supervised learning \cite{iccv2015Gupta_context,eccv2016zhang_color,eccv2016noroozi_jigsaw,iccv2017noroozi_count,cvpr2017zhang_splitbrain},
and 
{\bf(4)} Generative models 
\cite{nips2014goodfellow_gan,jmlr2010vincent_sae}.

{\bf Clustering analysis}
is a long-standing approach to unsupervised machine learning \cite{aggarwal2013data}. 
With the surge of deep learning techniques, recent studies have attempted to 
optimise clustering analysis and representation learning jointly
for maximising their complementary benefits
\cite{eccv2018Caron_deepclustering,icml2016xie_unsupervised,icml2016yang_kmeans-friendly,ICCV2017dizaji_clustering-autoencoder}.
Regardless, the key remains the discovery of
multiple class consistent clusters (or groups)
on the entire training data.
This is a difficult task
with the complexity and solution space exponentially proportional to
both the data and cluster size.
It is particularly so for clustering the data in complex structures and distributions
such as images and videos.
In contrast, the proposed AND model 
replaces the clustering operation with
local neighbourhood identification
in a {\em divide-and-conquer} principle.
This enables the control and mitigation of
the clustering errors and their negative propagation,
potentially yielding more accurate 
inference of latent class decision boundaries.

{\bf Sample specificity learning} goes to the other extreme
by considering every single sample as an independent class
\cite{cvpr2018wu_unsupervised,icml2017bojanowski_nat}.
The key idea is that supervised deep learning of neural networks
automatically reveals the visual similarity correlation 
between different classes from end-to-end optimisation.
However, this sort of supervision does not explicitly
model the class decision boundaries 
as clustering analysis and the AND model.
It is therefore likely to yield more
ambiguous class structures and less discriminative   
feature representations.

{\bf Self-supervised learning} has recently gained increasing research efforts
\cite{iccv2015Gupta_context,eccv2016zhang_color,eccv2016noroozi_jigsaw,iccv2017noroozi_count,cvpr2017zhang_splitbrain}.
Existing methods vary essentially
in the design of unsupervised auxiliary supervision.
Typically, such auxiliary supervision is hand-crafted 
to exploit some information intrinsically available in the
unlabelled training data, such as spatial context
\cite{iccv2015Gupta_context,eccv2016noroozi_jigsaw},
spatio-temporal continuity \cite{wang2015unsupervised,wang2017transitive},
and colour patterns \cite{eccv2016zhang_color,eccv2016larsson_color}. 
Due to the weak correlation with the underlying class targets, 
such learning methods mostly yield less discriminative models
than clustering analysis and our AND method.
How to design more target related auxiliary supervision remains an open problem.

\noindent {\bf Generative model} is a principled way of  
learning the true data distribution of the training set
in an unsupervised manner.
The most commonly used and efficient generative models 
include
Restricted Boltzmann Machines \cite{icml2009lee_conv-dbn,neural2006hinton_fast-dbn,cvpr2012tang_robust-RBM}, 
Autoencoders \cite{online2011Ng_sparse-autoencoder,jmlr2010vincent_sae}, and 
Generative Adversarial Networks 
\cite{iclr2016radford_dcgan,nips2014goodfellow_gan}.
The proposed AND model does not belong to this family,
but potentially generates complementary feature representations
due to a distinct modelling strategy.

Broadly, AND relates to constrained clustering
\cite{wagstaff2001constrained,kamvar2003spectral,zhu2013constrained,zhu2016constrained}
if considering our neighbourhood constraint as a form of
pairwise supervision including must-link and cannot-link.
However, our method is totally unsupervised without the 
need for pairwise links therefore more scalable.

\section{Unsupervised Neighbourhood Discovery}

Suppose we have $N$ training images 
$\mathcal{I} = \{\bm{I}_1, \bm{I}_2, ..., \bm{I}_N\}$. 
In unsupervised learning, no class labels are annotated on images.
The objective is to derive a deep CNN model $\bm{\theta}$ from 
the imagery data $\mathcal{I}$ that allows to extract 
class discriminative feature representations $\bm{x}$, 
$f_{\bm{\theta}}: \bm{I} \rightarrow \bm{x}$.
Without the access to class labels, it is unsupervised how the feature points
$\bm{x}$ should be distributed in training so that they can correctly represent 
the desired class memberships.
It is therefore necessary for an unsupervised learning algorithm 
to reveal such discriminative information directly from the visual data.
This is challenging due to the arbitrarily complex appearance patterns
and variations typically exhibited in the image collections
both within and across classes,
implying a high complexity of class decision boundaries.

\begin{figure}[th]
    \begin{center}
        \includegraphics[width=1.0\linewidth]{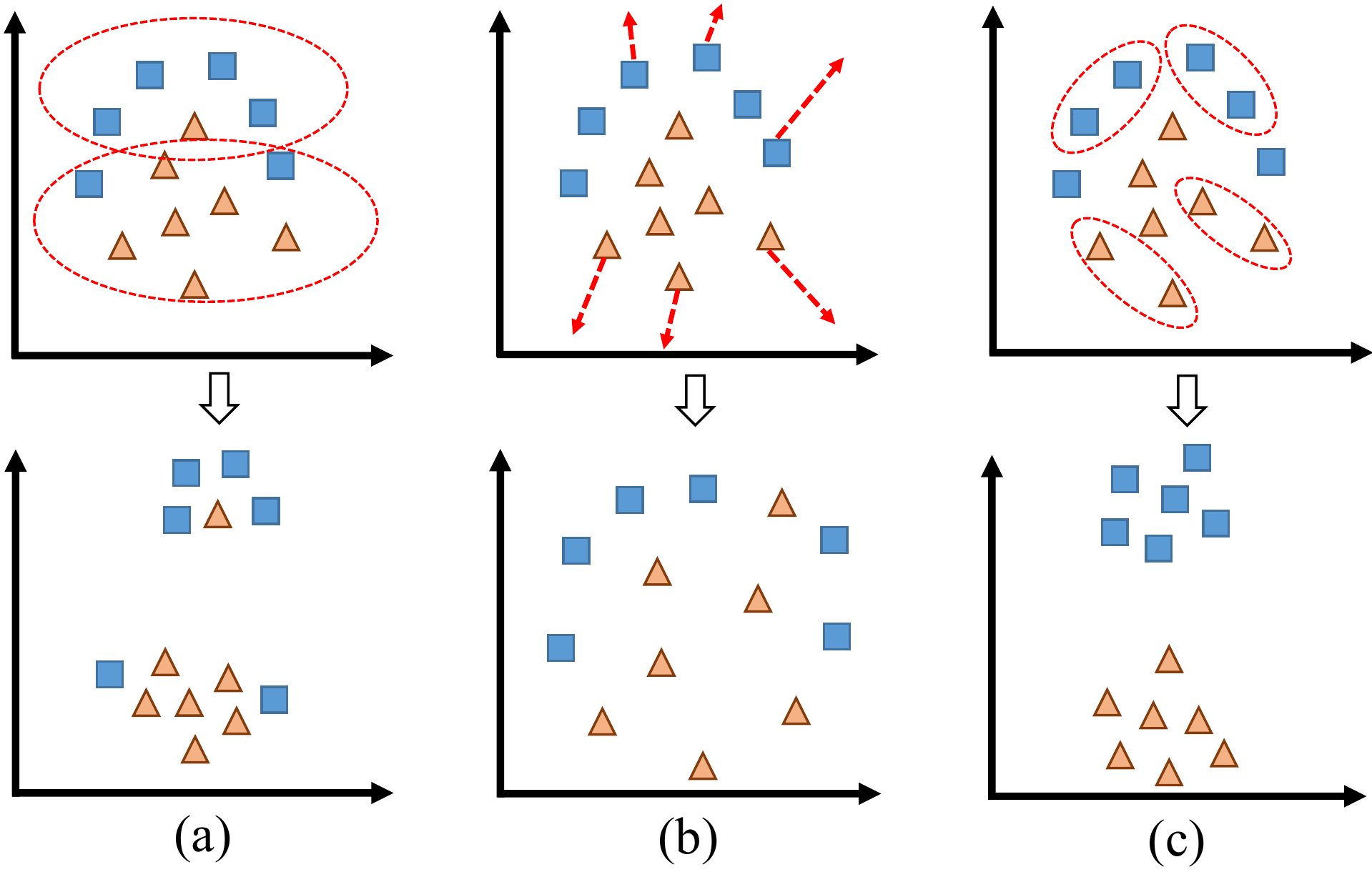}
    \end{center}
    \vskip -0.5cm
    \caption{
        Illustration of three unsupervised learning strategies.
        {\bf (a)} {\em Clustering analysis} aims for discovering the global class decision boundary \cite{eccv2018Caron_deepclustering,icml2016xie_unsupervised};
        {\bf (b)} {\em Sample specificity learning} discards the concept of clusters
        by treating every training sample as an independent class
         \cite{cvpr2018wu_unsupervised,icml2017bojanowski_nat};
        {\bf (c)} Our {\em Anchor Neighbourhood Discovery} searches local neighbourhoods
        with high class consistency.
    }
    \label{fig:ideas}
    \vspace{-0.3cm}
\end{figure}

To overcome the aforementioned problem,
we formulate an \textbf{\em Anchor Neighbourhood Discovery} (AND) method.
It takes a {\em divide-and-conquer} strategy from the local 
{sample anchored neighbourhood} perspective.
The key idea is that, whilst it is difficult and error-prone to
{directly} reason the {\em global class decision boundaries}
at the absence of class labels on the training data (Fig \ref{fig:ideas}(a)),
it would be easier and more reliable to estimate 
{\em local class relationship} in small neighbourhoods (Fig \ref{fig:ideas}(c)).  
Although such information is {\em incomplete} and provides
{\em less} learning supervision than the conventional clustering strategy
\cite{eccv2018Caron_deepclustering,icml2016xie_unsupervised} 
that operates at the coarse group level and mines
the clusters of data samples,
it favourably mitigates the misleading effect of noisy supervision.
%
Besides, the proposed AND model differs dramatically from 
the sample specificity learning strategy \cite{cvpr2018wu_unsupervised,icml2017bojanowski_nat}
that lacks a fundamental ability to mine the inter-sample class relationships
primitive to the global class boundaries (Fig \ref{fig:ideas}(b)).   
Therefore, 
the proposed method represents a conceptual trade-off 
between the two existing strategies
and a principled integration of them.

\begin{figure*}[t]
    \begin{center} 
        \includegraphics[width=1.0\linewidth]{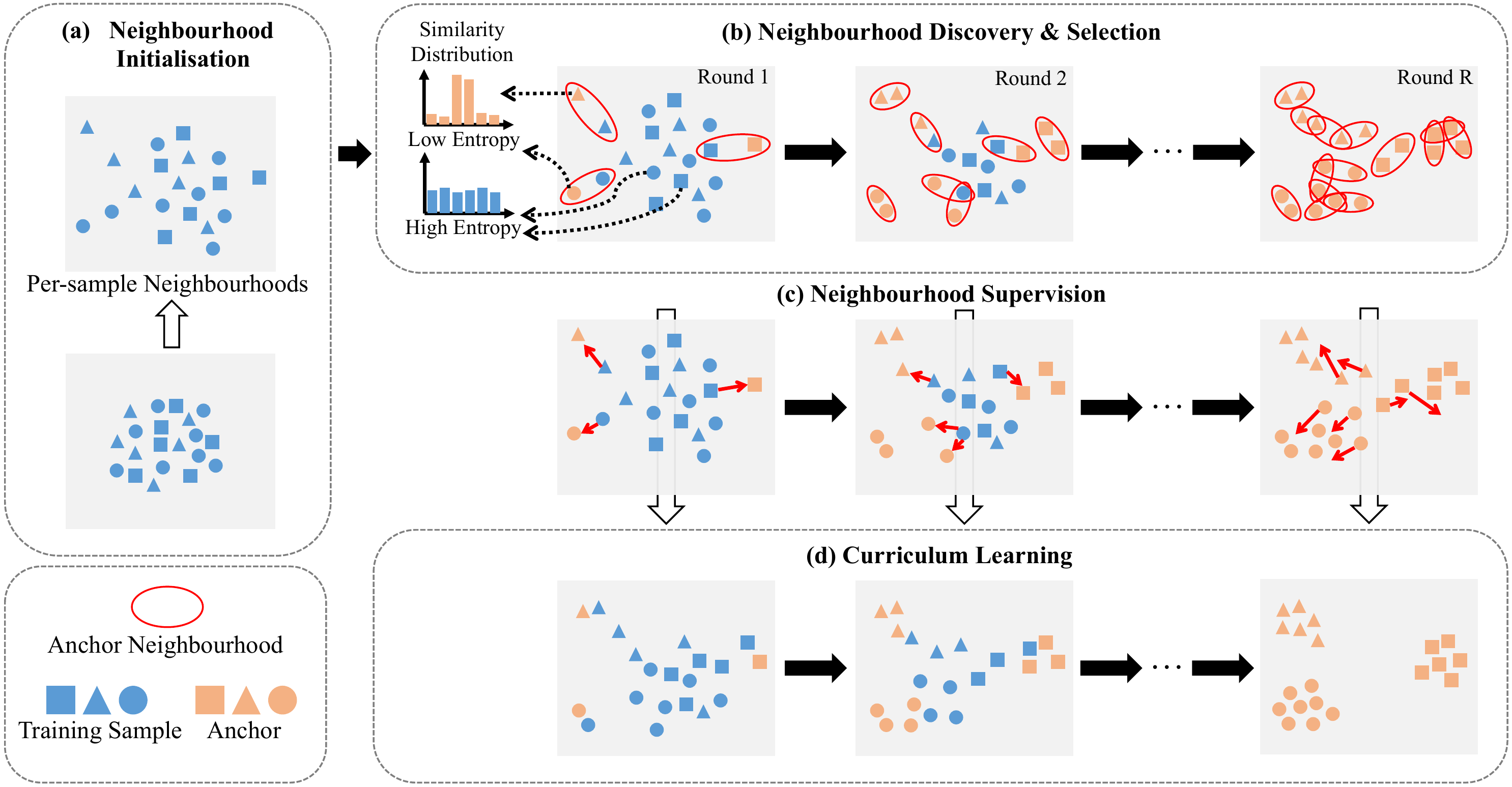}
    \end{center}
    \vskip -0.3cm
    \caption{
        Overview of the proposed {\em Anchor Neighbourhood Discovery} (AND) method
        for unsupervised deep learning.
        {\bf(a)} The AND model starts with per-sample neighbourhoods
        for model initialisation. 
        {\bf(b)} The resulting feature representations are then used to
        discover the local neighbourhoods anchored to every single
        training sample, i.e. anchor neighbourhoods.
        {\bf(c)} To incorporate the neighbourhood structure information
        into model learning, we propose a differentiable neighbourhood supervision 
        loss function for enabling end-to-end model optimisation.
        {\bf(d)} For enhancing model discriminative learning, 
        we further derive a curriculum learning algorithm for
        selecting class consistent neighbourhoods in a progressive manner.
        This is based on a novel similarity distribution entropy measurement.
    }
    \label{fig:pipeline}
    \vspace{-0.3cm}
\end{figure*}

As shown in our evaluations, 
the proposed training strategy yields
superior models. 
This indicates the significance of both minimising 
the erroneous self-mined supervision and exploiting the inter-sample
class relations spontaneously during unsupervised learning.
An overview of the proposed AND model is depicted in
Fig \ref{fig:pipeline}.

\subsection{Neighbourhood Discovery}
We start with how to identify neighbourhoods.
An intuitive method is using $k$ nearest neighbours ($k$NN)
given a feature space $X$ and a similarity metric $s$, 
e.g. the cosine similarity (Fig \ref{fig:pipeline}(b)).
A neighbourhood $\mathcal{N}_k(\bm{x})$ determined by $k$NN is sample-wise, i.e. anchored to
a specific training sample $\bm{x}$:
\begin{equation}
\mathcal{N}_k(\bm{x}) = \{\bm{x}_i \;\; | \;\; s(\bm{x}_i, \bm{x}) {\text{ is top-}k} \text{ in } X\} \cup \{\bm{x}\}, 
\label{eq:KNN}
\end{equation}
where $X$ denotes the feature space.
We call such structures as \textbf{\em Anchor Neighbourhoods} (\texttt{AN}).

To enable class discriminative learning,
we want all samples in a single neighbourhood \texttt{AN} to share the same class label,
i.e. {\em class consistent}.
As such, we can facilitate 
the design of learning supervision
by assigning the same label to these samples.
This requirement, however, is non-trivial to fulfil in
unsupervised learning since we have no reasonably good
sample features,
even though a neighbourhood \texttt{AN} can be much smaller and more
local (therefore likely more class consistent) than a typical cluster
when using small $k$ values.
Moreover, we begin with the training images but {\em no} learned features.
%
This even prevents the formation of $\mathcal{N}_k$ and 
gives rise to an extreme case -- 
each individual sample represents a distinct 
anchor neighbourhood.

\noindent{\bf Neighbourhood Initialisation.}
Interestingly, such initial \texttt{AN}s
are in a similar spirit of sample specificity learning \cite{cvpr2018wu_unsupervised,icml2017bojanowski_nat}
where each data instance is assumed to represent a distinct class
(Fig \ref{fig:pipeline}(a)).  
With this conceptual linkage, we exploit the instance loss 
\cite{cvpr2018wu_unsupervised} to commence the model learning.
Specifically, it is a non-parametric variant of the 
softmax cross-entropy loss written as:
\begin{equation}
\mathcal{L}_{\text{init}} = - \sum_{i=1}^{n_\text{bs}} \log(p_{i,i}), \;
p_{i,j} = \frac{\text{exp}(\bm{x}_i^\top \bm{x}_j / \tau)}{\sum_{k=1}^N \text{exp}(\bm{x}_i^\top \bm{x}_k / \tau)} 
\label{eq:predictions}
\end{equation}
where $n_\text{bs}$ denotes the training mini-batch size,
and the temperature parameter $\tau$ is
for controlling the distribution concentration degree~\cite{arxiv2015hinton_distill}.


\noindent{\bf Neighbourhood Supervision.}
In the feature space derived by Eq \eqref{eq:predictions},
we build a neighbourhood $\mathcal{N}_k(\bm{x})$ for each individual sample $\bm{x}$.
Considering the high appearance similarity among the samples of each $\mathcal{N}_k(\bm{x})$,
we assume they share a single class label 
for model discriminative learning.

Formally, we formulate an unsupervised neighbourhood supervision signal 
as:
\begin{equation}
\mathcal{L}_\text{AN} = -\sum_{i=1}^{n_\text{bs}}
\log \Big( \sum_{j\in \mathcal{N}_k(\bm{x}_i)} p_{i,j} \Big)
\label{eq:AN_loss}
\end{equation}

The rationale behind Eq \eqref{eq:AN_loss} 
is to encourage label consistency for anchor neighbourhoods
(Fig \ref{fig:pipeline}(c)).
Specifically, the probability $p_{i,j}$ (Eq \eqref{eq:predictions}), obtained using a softmax function, represents 
visual similarity between $\bm{x}_i$ and $\bm{x}_j$
in a stochastic manner.
This takes the spirit of {\em stochastic nearest neighbour}
\cite{nips2005goldberger_nca},
as it considers the entire training set. 
In this scheme, 
the probability $p(\bm{x})$ of correctly classifying a sample $\bm{x}_i$ 
can be then represented as:
\begin{equation}
p(\bm{x}_i) =
\sum_{j \in C} p_{i,j}
\label{eq:SNN_prob}
\end{equation}
where $C$ denotes the set of samples in the same class as $\bm{x}_i$.
However, $C$ is unavailable to unsupervised learning.
To overcome this problem,
we approximate $C$ by the neighbourhoods \texttt{AN}s,
each of which is likely to be class consistent.
Together with the cross-entropy function,
this finally leads to the formulation of the proposed $\mathcal{L}_\text{AN}$ loss
(Eq \eqref{eq:AN_loss}).

\textbf{\em Remarks.}
The proposed neighbourhood supervision formulation $\mathcal{L}_{\text{AN}}$ 
aims at exploring the {\em local class information},
under the assumption that anchor neighbourhoods are class consistent.
This is because each neighbourhood \texttt{AN} is treated 
as a different learning concept (e.g. class),
although some \texttt{AN}s may share the same unknown class label.
Such information is also {\em partial} due to that
a specific \texttt{AN} may represent only a small  
proportion of the corresponding class,
and multiple \texttt{AN}s with the same underlying class can represent
different aspects of the same concept {\em collectively} 
(not the whole view due to no \texttt{AN}-to-\texttt{AN} relationships).
It is the set of these distributed anchor neighbourhoods {\em as a whole}
that brings about the class discrimination capability
during model training.
It is in a {\em divide-and-conquer} principle.

Fundamentally, the proposed design differs dramatically from both (1)
{\em the clustering strategy} that seeks for the complete 
class boundary information -- a highly risky and error-prone process
\cite{eccv2018Caron_deepclustering,icml2016xie_unsupervised},
and 
(2)
the {\em sample specificity learning} that instead totally ignores the class level
information therefore less discriminative
\cite{cvpr2018wu_unsupervised,icml2017bojanowski_nat}.
Moreover, clustering often requires the prior knowledge
of the cluster number therefore limiting their usability 
and scalability due to the lack of it in many applications. 
On the contrary, this kind of information is not needed for forming 
the proposed \texttt{AN}s, therefore more application generic and scalable.
To maximise the class consistency degree in \texttt{AN}s, 
we simply need to use the smallest neighbourhood size,
i.e. $k\!=\!1$.



{\bf Neighbourhood Selection.}
As discussed above, the proposed method requires the neighbourhoods
\texttt{AN}s to be class consistent.
This condition, nonetheless, is difficult to meet. 
Specifically, the instance loss function 
$\mathcal{L}_{\text{init}}$ (Eq \eqref{eq:predictions}) 
encourages the feature representation learning towards that
each sample's specificity degree can be maximised as possible
on the training data.
Considering a sample $\bm{x}_i$,
other samples either share the class label (positive) with $\bm{x}_i$
or not (negative).
Hence, this formulation may yield a model with certain discrimination ability,
e.g. when a subset of (unknown) positive samples are associated
with similar visual specificity.   
But this entirely depends on the intrinsic data properties 
without stable guarantee.
It means that typically {\em not} all neighbourhoods \texttt{AN}s
are reliable and class consistent.
This inevitably leads to the necessity of
conducting neighbourhood selection for more reliable 
model learning.

To this end, we go beyond by 
taking advantages of the curriculum learning idea
\cite{icml2009bengio_curriculum, dong2017multi}.
Instead of taking a one-off neighbourhood selection,
we introduce a {\em progressive} selection process (Fig \ref{fig:pipeline}(d))
which distributes evenly the neighbourhood selection 
across $R$ rounds.
This realises an easy-to-hard learning procedure through a curriculum.



\underline{\em Selecting Curriculum.}
To enable automated neighbourhood selection for 
making a scalable curriculum, it is necessary for us
to derive a selecting criterion.
This is achieved by exploiting the intrinsic nature
of the probability $p_{i,j}$ (Eq \eqref{eq:predictions})
defined between two samples $\bm{x}_i$ and $\bm{x}_j$.
More specifically, we utilise the entropy measurement of 
the probability vector $p_i = [p_{i,1}, p_{i,2}, \cdots, p_{i,N}]$
as the class consistency indicator of the corresponding neighbourhood \texttt{AN}
as:
\begin{equation}
H(\bm{x}_i) = - \sum_{j=1}^{N} p_{i,j}\log(p_{i,j}).
\end{equation}
We consider that smaller $H(\bm{x}_i)$ values correspond to
more consistent neighbourhoods. 
In particular, when $H(\bm{x}_i)$ is small,
it means $\bm{x}_i$ resides in a low-density area with sparse visual similar
neighbours surrounding.
In the definition of sample specificity learning (Eq \eqref{eq:predictions}),
the model training tends to converge to some local optimum
that all samples of a neighbourhood $\mathcal{N}_k(\bm{x}_i)$ with small $H(\bm{x}_i)$ 
share some easy-to-locate visual appearance, and simultaneously the same underlying class label statistically since positive samples are more likely to present such appearance commonness including the context than negative ones.
On the contrary, 
a large $H(\bm{x}_i)$ implies a neighbourhood $\mathcal{N}_k(\bm{x}_i)$ 
residing in a dense area, a case that
the model fails to identify the sample specificity.
This is considered hard cases,
and requires more information for the model to interpret them.

In light of the observations above,
we formulate a linear curriculum according to the 
class consistency entropy measurement.
Specifically, for the $r$-th round (among a total of $R$ rounds),
we select the top-$S$ (Eq \eqref{eq:curri}) of \texttt{AN}s according to their corresponding entropy for model learning
by the proposed neighbourhood supervision loss $\mathcal{L}_{\text{AN}}$ (Eq \eqref{eq:AN_loss}).
\begin{equation}
S = \frac{r}{R}*100\%
\label{eq:curri}
\end{equation}
Since the remaining training samples are still not sufficiently interpreted 
by the model at the current round, they are preserved as individuals 
(i.e. single-sample neighbourhoods) 
as in sample specificity learning (Eq \eqref{eq:predictions}).

\textbf{Objective Loss Function.}
With the progressive neighbourhood discovery as above,
we obtain the model objective loss function for the $r$-th round
as: 
\begin{equation}
\mathcal{L}^r = - \sum_{i \in B^r_\text{inst}} \log (p_{i,i})
- \sum_{i \in B^r_\text{AN}} \log \Big( \sum_{j \in \mathcal{N}_k(\bm{x}_i)} p_{i,j}\Big)
\label{eq:loss_round}
\end{equation}
where $B^r_\text{inst}$ and $B^r_\text{AN}$
denote the set of instances and the set of
\texttt{AN}s in a mini-batch at the $r$-th round, respectively.

As each round of training is supposed to 
improve the model, we update the neighbourhoods \texttt{AN}s
for all training samples before 
performing neighbourhood selection per round.
To facilitate this process,
we maintain an offline memory to store the feature vectors.
We update the memory features of mini-batch samples 
by exponential moving average~\cite{lucas1990exponentially}
over the training iterations as:
\begin{equation}
\tilde{\bm{x}}_i = (1 - \eta) \cdot \tilde{\bm{x}}_i + \eta \cdot \bm{x}_i
\label{eq:feat_update}
\end{equation}

where 
$\eta$ denotes the update momentum,
${\bm{x}}_i$ and $\tilde{\bm{x}}_i$ the up-to-date 
and memory feature vector respectively.

\subsection{Model Optimisation}

The proposed loss function (Eq \eqref{eq:loss_round}) is differentiable
therefore enabling the stochastic gradient descent algorithm
for model training.
In particular, 
when $\bm{x}_i$ comes as an individual instance,
the gradients for $\mathcal{L}^r$ w.r.t. $\bm{x}_i$
and $\bm{x}_j$ ($j \! \neq \!i$) are written as:
\begin{equation}\small
\frac{\partial \mathcal{L}^r}{\partial \bm{x}_i} = 
\frac{1}{\tau}[\sum_{k=1}^N (p_{i,k} \cdot \bm{x}_k) + (p_{i,i} - 2) \cdot \bm{x}_i ], \;\;
\frac{\partial \mathcal{L}^r}{\partial \bm{x}_j} = \frac{1}{\tau}p_{i,j} \cdot \bm{x}_i
\label{eq:grad_inst}
\end{equation}
When $\bm{x}_i$ corresponds to an \texttt{AN},
the gradients are then:
\begin{equation}\small
\frac{\partial \mathcal{L}^r}{\partial \bm{x}_i} = 
\frac{1}{\tau}[\sum_{k=1}^N (p_{i,k} \cdot \bm{x}_i) - 
\sum_{k \in \mathcal{N}_k(\bm{x}_i)} 
\tilde{p}_{i,k} + (p_{i,i} - \tilde{p}_{i,i}) \cdot \bm{x}_i]
\label{eq:grad_AN_i}
\end{equation}
\begin{equation}\label{eq:grad_AN_j}
\small
\frac{\partial \mathcal{L}^r}{\partial \bm{x}_j} = \left\{
\begin{array}{lcl}
\frac{1}{\tau}[p_{i,j} \cdot \bm{x}_i - \tilde{p}_{i,j} \cdot \bm{x}_i], & & j \in \mathcal{N}_k(\bm{x}_i) \\
\frac{1}{\tau}[p_{i,j} \cdot \bm{x}_i], & & j \notin \mathcal{N}_k(\bm{x}_i)
\end{array} \right.
\end{equation}
where $\tilde{p}_{i,j} = p_{i,j}/\sum_{k \in \mathcal{N}_k(\bm{x}_i)}p_{i,k}$ is the normalised distribution over the neighbours. 
The whole model training procedure is summarised in Algorithm
\ref{Algorithm}. 

\begin{algorithm}[h]
    \caption{Neighbourhood discovery.} \label{Algorithm}
    \textbf{Input:} 
    Training data $\mathcal{I}$,
    rounds $R$, iterations per round $T$. \\
    \textbf{Output:}
    A deep CNN feature model. \\
    \textbf{Initialisation:} 
    Instance specificity learning (Eq \eqref{eq:predictions}). \\
    \textbf{Unsupervised learning:} \\
    \textbf{for} $r=1$ \textbf{to}  \textsl{R} \textbf{do} \\ 
    \hphantom{~~}
    Form neighbourhoods with the current features (Eq \eqref{eq:KNN});\\
    \hphantom{~~}
    Curriculum selection of neighbourhoods (Eq \eqref{eq:curri});\\
    \hphantom{~~}
    \textbf{for} $t=1$ \textbf{to} \textsl{$T$} \textbf{do} \\ 
    \hphantom{~~~~} 
    Network forward propagation (batch feed-forward);\\
    \hphantom{~~~~} 
    Objective loss computation (Eq~\eqref{eq:loss_round});\\
    \hphantom{~~~~} 
    Network back-propagation (Eq~\eqref{eq:grad_inst},\eqref{eq:grad_AN_i},\eqref{eq:grad_AN_j});\\
    \hphantom{~~~~} 
    Memory feature update (Eq~\eqref{eq:feat_update}).\\
    \hphantom{~~}
    \textbf{end for}\\
    \textbf{end for} 
\end{algorithm}

\section{Experiments}
\label{sec:exps}

\begin{figure}[h]
    \begin{center}
        \includegraphics[width=1.0\linewidth]{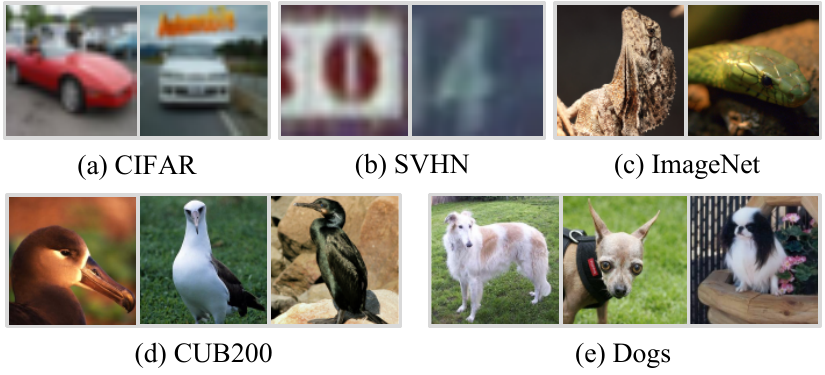}
    \end{center}
    \vskip -0.5cm
    \caption{
        Dataset example images. 
    }
    \label{fig:datasets}
    \vspace{-0.5cm}
\end{figure}

\noindent{\bf Datasets.}
We used 6 image classification benchmarks for
evaluating our model (Fig \ref{fig:datasets}).
\textbf{\em CIFAR10(/100)}~\cite{krizhevsky2009cifar}: 
An image dataset with 50,000/10,000 train/test images from 10 (/100) object classes. 
Each class has 6,000 (/600) images with size $32\!\times\!32$.
%
\textbf{\em SVHN}~\cite{netzer2011svhn}: 
A Street View House Numbers dataset including 10 classes of digit images.
%
\textbf{\em ImageNet}~\cite{russakovsky2015imagenet}: 
A large 1,000 classes object dataset with 1.2 million images for training and 50,000 for test.
%
\textbf{\em CUB200-2011}~\cite{Wah2011CUB200}: 
A fine-grained dataset containing 5,994/5,794 train/test images of 200 bird species.
%
\textbf{\em Stanford Dogs}~\cite{khosla2011dogs}: 
A fine-grained dataset 
with 12,000/8,580 train/test images of 120 dog breeds.

\noindent{\bf Experimental setup.}
For learning any unsupervised representation model, 
we assumed and used only the training image data but {\em no} class labels.
Unless stated otherwise, we adopted the AlexNet \cite{nips2012krizhevsky_CNN} 
as the neural network architecture
for fair comparisons with the state-of-the-art 
methods.
To assess the quality of a learned model for classification at test time,
we utilised the ground-truth class labels of the training images
{\em merely} for enabling image categorisation.
This does not change the feature representations
derived in unsupervised learning.

Following \citet{cvpr2018wu_unsupervised}, 
we considered two classification models,
{\em Linear Classifier} (LC),
and
{\em Weighted $k$NN}, 
as well as the feature representations 
extracted from different network layers respectively.
LC was realised by a fully connected (FC) layer
optimised by the cross-entropy loss function.
The non-parametric classifier $k$NN predicts the class label
by weighted voting of top-$k$ neighbours $\mathcal{N}_k$ as 
$s_c  = \sum_{i\in \mathcal{N}_k} \delta(c,c_i) \cdot w_i$
where $\delta(c,c_i)$ is the Dirac delta function which returns $1$ if $c=c_i$,
and 0 otherwise.
The weight $w_i$ is computed from the cosine similarity $s_i$ as
$w_i = \exp(s_i/\tau)$ with $\tau=0.07$ the temperature parameter.
Without an extra classifier learning post-process involved,
$k$NN reflects {\em directly} 
the discriminative capability of the learned feature representations.

\noindent {\bf Performance metric.}
We adopted the top-1 classification accuracy for the model performance
measurement.

\noindent {\bf Competitors.}
We compared the proposed AND model with four
types of state-of-the-art unsupervised learning methods:
{\bf(1)} {\em Generative model}: BiGAN~\cite{iclr2016donahue_bigan};
{\bf(2)} 
{\em Clustering method}: 
DeepCluster~\cite{eccv2018Caron_deepclustering};
{\bf(3)} 
{\em Self-supervised learning methods}: 
Context~\cite{iccv2015Gupta_context}, 
Colour~\cite{eccv2016zhang_color}, 
Jigsaw~\cite{eccv2016noroozi_jigsaw}, 
Counting~\cite{iccv2017noroozi_count},
and Split-Brain~\cite{cvpr2017zhang_splitbrain};
{\bf (4)}
{\em Sample specificity learning methods}:
Instance~\cite{cvpr2018wu_unsupervised} and 
Noise As Targets (NAT)~\cite{icml2017bojanowski_nat}; in total 9 methods.

\noindent{\bf Implementation details.}
For fair comparisons, we used the same experimental setting as \cite{cvpr2018wu_unsupervised,icml2017bojanowski_nat}.
To train AND models, 
we set the learning rate to 0.03 which
was further scaled down by 0.1 every 40 epochs after the first 80 epochs.
We used the batch size of 256 for ImageNet and 128 for others.
We set the epoch to 200 per round.
We fixed the feature length to 128.
We applied the SGD with Nesterov momentum at 0.9.
Our model usually converges with $R\!=\!4$ rounds.
We set $\eta\!=\!0.5$ in Eq \eqref{eq:feat_update} for feature update.
We set $k\!=\!1$ (Eq \eqref{eq:KNN}) for exploring the most local neighbourhoods.

\subsection{Comparisons to the State-of-the-Art Methods}
\label{ssec:comparsion}

{\bf Small scale evaluation.}
Table \ref{tab:state-of-the-art_small-scaled}
compares the object image classification results 
on three benchmarks
between AND and four unsupervised learning methods.
We tested two classification models,
weighted $k$NN using FC features
and 
linear regression using conv5 features.
We have these observations:
{\bf(1)} The AND method performs best often by large margins over all competitors, 
except linear regression on CIFAR10 with DeepCluster outperforms marginally. 
This suggests the performance superiority of our neighbourhood
discovery over alternative methods
for unsupervised representation learning.
{\bf(2)} The margins obtained by $k$NN tend to be larger 
than those by linear regression.
This indicates that AND features 
are favourably more ready for {\em direct} use
without extra classifier training as post-processing.

\begin{table} [h]
\begin{center}
\begin{tabular}{|l|c|c|c|}
\hline
\cline{2-4}
Dataset & CIFAR10 & CIFAR100 & SVHN \\ 
\hline \hline
Classifier/Feature & \multicolumn{3}{c|}{Weighted $k$NN / FC} \\
\hline
Split-Brain$^*$
& 11.7 & 1.3 & 19.7 \\
Counting$^*$
& 41.7 & 15.9 & 43.4 \\
DeepCluster 
& \second{62.3} & 22.7 & \second{84.9} \\
Instance 
& 60.3 & \second{32.7} & 79.8 \\
\hline
\bf AND 
& \best{74.8} & \best{41.5} & \best{90.9} \\
\hline
\em Supervised
& 91.9 & 69.7 & 96.5 \\
\hline \hline
Classifier/Feature & \multicolumn{3}{c|}{Linear Classifier / conv5} \\ \hline
Split-Brain$^*$
& 67.1 & 39.0 & 77.3 \\
Counting$^*$
& 50.9 & 18.2 & 63.4 \\
DeepCluster 
& \best{77.9} & \second{41.9} & \second{92.0} \\
Instance 
& 70.1 & 39.4 & 89.3 \\
\hline
\bf AND 
&\second{77.6} & \best{47.9} & \best{93.7} \\
\hline
\em Supervised
& 91.8 & 71.0 & 96.1 \\
\hline 
\end{tabular}
\end{center}
\vskip -0.3cm
\caption{
    Evaluation on small scale image datasets.
    $^*$: By a reproduced implementation.
}
\vspace{-0.5cm}
\label{tab:state-of-the-art_small-scaled}
\end{table}

{\bf Large scale evaluation.}
We evaluated the scalability of our AND model
using ImageNet.
Table \ref{tab:state-of-the-art} compares AND with nine alternative 
methods.
Following the previous studies, we tested all conv layers.
The results show that:
{\bf (1)} All unsupervised learning methods clearly
surpass the random features, suggesting their modelling effectiveness
consistently.
{\bf (2)} AND outperforms 
all competitors but by smaller margins. 
This is likely due to using over tiny neighbourhoods (sized 2) 
for being consistent with small scale datasets.
The amount of inter-sample relationships is quadratic to the data size,
so bigger neighbourhoods may be beneficial for large datasets
in capturing structural information. 
{\bf (3)} 
Most unsupervised learning methods
yield the respective best representation
at intermediate layers when using linear classifier. 
The plausible reason is that their supervision singles are less
correlated with the ground-truth targets.


\begin{table} [h]\footnotesize
    \setlength{\tabcolsep}{0.07cm}
    \begin{center}
        \begin{tabular}{|l|c|c|c|c|c|c|c|}
            \hline
            Classifier & \multicolumn{6}{c|}{Linear Classifier} & $k$NN \\ 
            \cline{1-8}
            Feature & conv1 & conv2 & conv3 & conv4 & conv5 & FC & FC \\
            \hline
            \hline
            \em Random & 11.6 & 17.1 & 16.9 & 16.3 & 14.1 & 12.0 & 3.5 \\
            \hline
            \em Supervised & 19.3 & 36.3 & 44.2 & 48.3 & 50.5 & - & - \\
            \hline \hline
            Context 
            & 17.5 & 23.0 & 24.5 & 23.2 & 20.6 & 30.4 & - \\
            BiGAN 
            & 17.7 & 24.5 & 31.0 & 29.9 & 28.0 & 32.2 & - \\
            Colour 
            & 13.1 & 24.8 & 31.0 & 32.6 & 31.8 & 35.2 & - \\
            Jigsaw 
            & \best{19.2} & 30.1 & 34.7 & 33.9 & 28.3 & \best{38.1} & - \\
            NAT 
            & - & - & - & - & - & 36.0 & - \\
            Counting 
            & \second{18.0} & \second{30.6} & 34.3 & 32.5 & 25.7 & - & -\\
            Split-Brain 
            & 17.7 & 29.3 & 35.4 & 35.2 & 32.8 & - & 11.8 \\
            DeepCluster 
            & 13.4 & \best{32.3} & \best{41.0} & \second{39.6} & \best{38.2} & - & \second{26.8} \\
            Instance 
            & 16.8 & 26.5 & 31.8 & 34.1 & 35.6 & - & \best{31.3} \\
            \hline
            \bf AND
            & 15.6 & 27.0 & \second{35.9} & \best{39.7} & \second{37.9} & \second{36.7} & \best{31.3} \\
            \hline
        \end{tabular}
    \end{center}
    \vskip -0.3cm
    \caption{
        Evaluation on ImageNet. 
        The results of existing methods 
        are adopted from~\cite{cvpr2018wu_unsupervised,icml2017bojanowski_nat}.
    }
    \vspace{-0.3cm}
    \label{tab:state-of-the-art}
\end{table}

{\bf Fine-grained evaluation.}
We evaluated AND with more challenging fine-grained recognition
tasks which are significantly under-studied in unsupervised learning context.
Consistent with the results discussed above, 
Table~\ref{tab:fine-grained} demonstrates again the
performance superiority of
our neighbourhood discovery idea
over the best competitor Instance. 

\begin{table} [h]
    \setlength{\tabcolsep}{0.7cm}
    \begin{center}
        \begin{tabular}{|l|c|c|}
            \hline
            Dataset & CUB200 & Dogs \\
            \hline\hline
            Instance & 11.6 & 27.0 \\
            \hline
            \bf AND & \bf{14.4} & \bf{32.3} \\
            \hline
        \end{tabular}
    \end{center}
    \vskip -0.3cm
    \caption{
        Evaluation on fine-grained datasets.
        Network: ResNet18.
    }
    \label{tab:fine-grained}
    \vspace{-0.5cm}
\end{table}

\subsection{Component Analysis and Discussions}
We conducted detailed component analysis
with the weighted $k$NN classifier and FC features.

{\bf Backbone network.}
We tested the generalisation of AND with varying-capacity networks.
We further evaluated ResNet18 and ResNet101 \cite{cvpr2016he_resnet}
on CIFAR10. 
Table \ref{tab:nets} shows that 
AND benefits from stronger net architectures.
A similar observation was made on 
ImageNet: 41.2\% (ResNet18) vs. 31.3\% (AlexNet).

\begin{table} [h]
    \begin{center}
        \begin{tabular}{|l|c|c|c|}
            \hline
            Network & AlexNet & ResNet18 & ResNet101 \\
            \hline\hline
            Accuracy & 74.8 & 86.3 & \bf{88.4} \\
            \hline
        \end{tabular}
    \end{center}
    \vskip -0.3cm
    \caption{
        Network generalisation analysis on CIFAR10.
    }
    \label{tab:nets}
    \vspace{-0.3cm}
\end{table}

{\bf Model initialisation.}
We tested the impact of initial features for neighbourhood discovery
by comparing random and Instance networks.  
Table~\ref{tab:initialisation} shows that AND can benefit
from stronger initialisation whilst being robust to weak initial representations.

\begin{table} [h]
    \begin{center}
        \begin{tabular}{|l|c|c|}
            \hline
            Initialisation & Random Model & Instance Model \\
            \hline\hline
            Accuracy  & 85.7 & \bf{86.3} \\
            \hline
        \end{tabular}
    \end{center}
    \vskip -0.3cm
    \caption{
        Effect of model initialisation  
        on CIFAR10.
    }
    \label{tab:initialisation}
    \vspace{-0.3cm}
\end{table}

\begin{figure}[h]
    \begin{center}
        \includegraphics[width=0.9\linewidth]{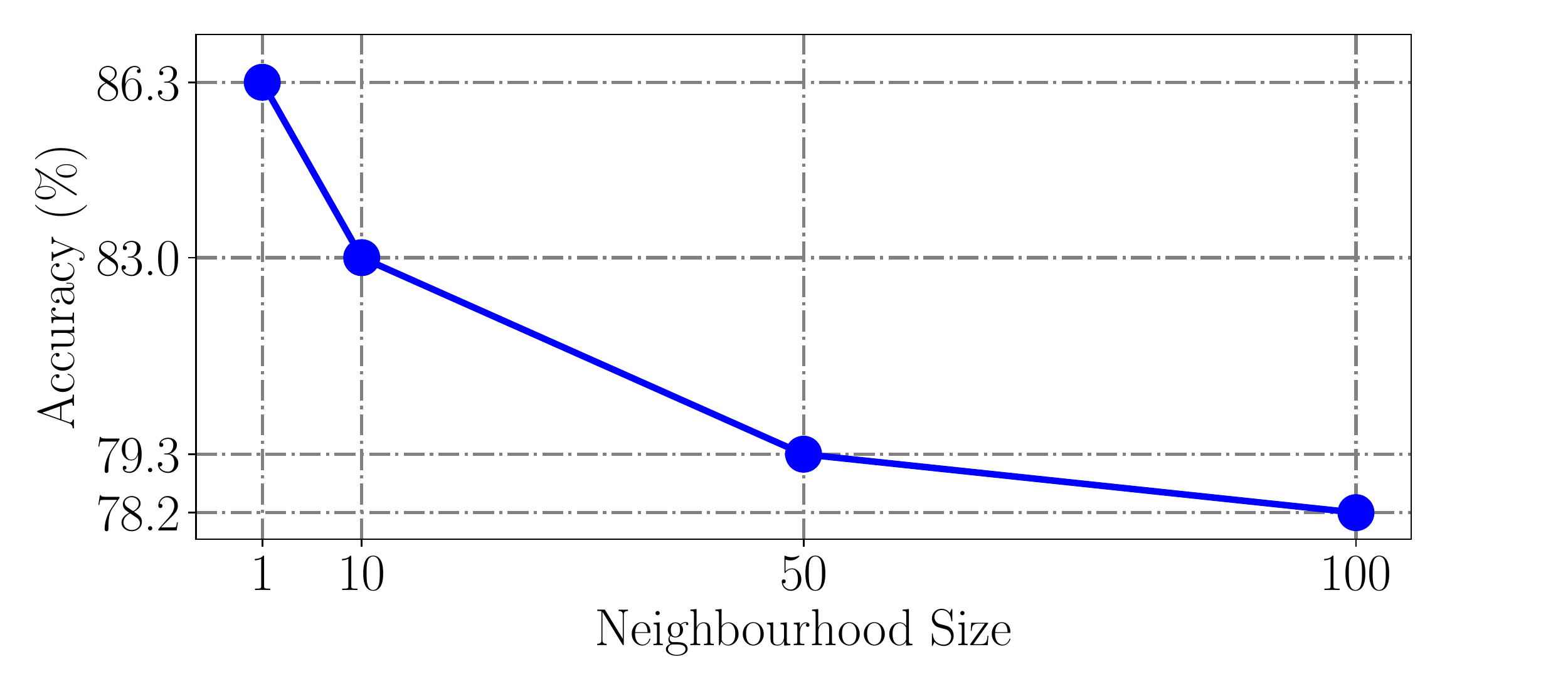}
    \end{center}
    \vskip -0.5cm
    \caption{Effect of the neighbourhood size on CIFAR10.}
    \label{fig:neighbour_size}
    \vspace{-0.3cm}
\end{figure}

{\bf Neighbourhood size.}
Neighbourhood size is an important parameter 
since it controls label consistency of \texttt{AN}s
and finally the model performance.
We evaluated its effect using ResNet18 on CIFAR10
by varying $k$ from 1 (the default value) to 100. 
Figure \ref{fig:neighbour_size} shows that the smallest neighbourhoods (i.e. $k\!=\!1$) 
are the best choice. 
This implies high variety of imagery data,
so smaller \texttt{AN}s are preferred  
for unsupervised learning.
%

\begin{figure}[h]
    \begin{center}
        \includegraphics[width=0.9\linewidth]{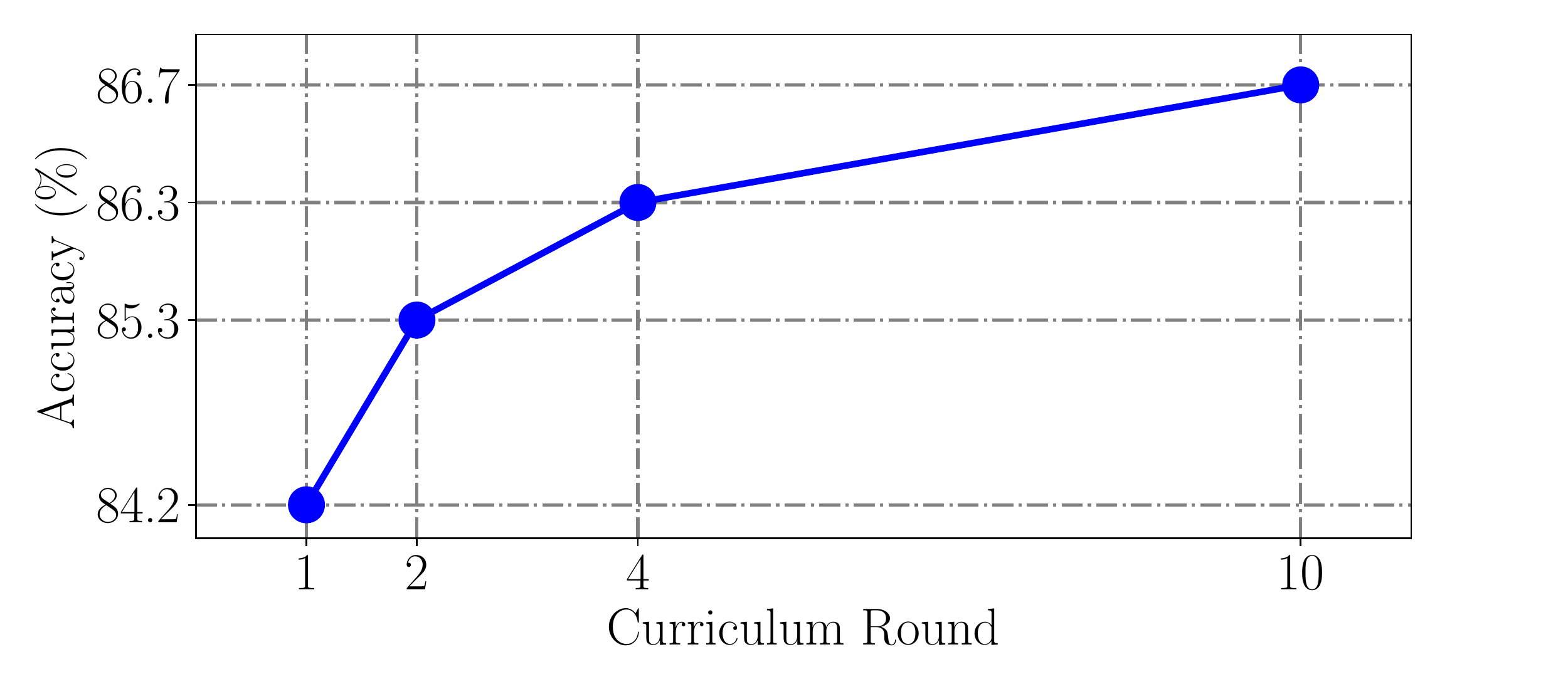}
    \end{center}
    \vskip -0.5cm
    \caption{Effect of the curriculum round on CIFAR10.}
    \vspace{-0.5cm}
    \label{fig:rounds}
\end{figure}

{\bf Curriculum round.}
We tested the effect of the curriculum round ($R$ in Eq \eqref{eq:curri}) 
of progressive neighbourhood discovery.
More rounds consume higher training costs.
Figure \ref{fig:rounds} shows that
using 4 rounds gives a good trade-off between model
training efficiency and feature performance. 
Often, per-round epoch number $N_\text{ep}$ affects
the training efficiency and performance.
We investigated its effect and
found that AND achieves 83.3\% 
by $N_\text{ep}$=50
vs. 84.8\% by $N_\text{ep}$=100.

{\bf One-off {\em vs.} curriculum neighbourhood discovery.}
We evaluated the benefit of AND's curriculum.
To this end, we compared with the {\em one-off} discovery counterpart
where all anchor neighbourhoods are exploited one time.
Table \ref{tab:oneoff} shows that 
the proposed multi-round progressive discovery via a curriculum 
is effective to discover more reliable anchor neighbourhoods
for superior unsupervised learning.

\begin{table} [h]
    \setlength{\tabcolsep}{0.5cm}
    \begin{center}
        \begin{tabular}{|l|c|c|}
            \hline
            Discovery & One-off & Curriculum  \\
            \hline\hline
            Accuracy  & 84.2 & \bf{86.3} \\
            \hline
        \end{tabular}
    \end{center}
    \vskip -0.3cm
    \caption{
        One-off vs. curriculum discovery on CIFAR10.
    }
    \vspace{-0.3cm}
    \label{tab:oneoff}
\end{table}

\begin{figure}[h]
    \begin{center}
        \includegraphics[width=0.9\linewidth]{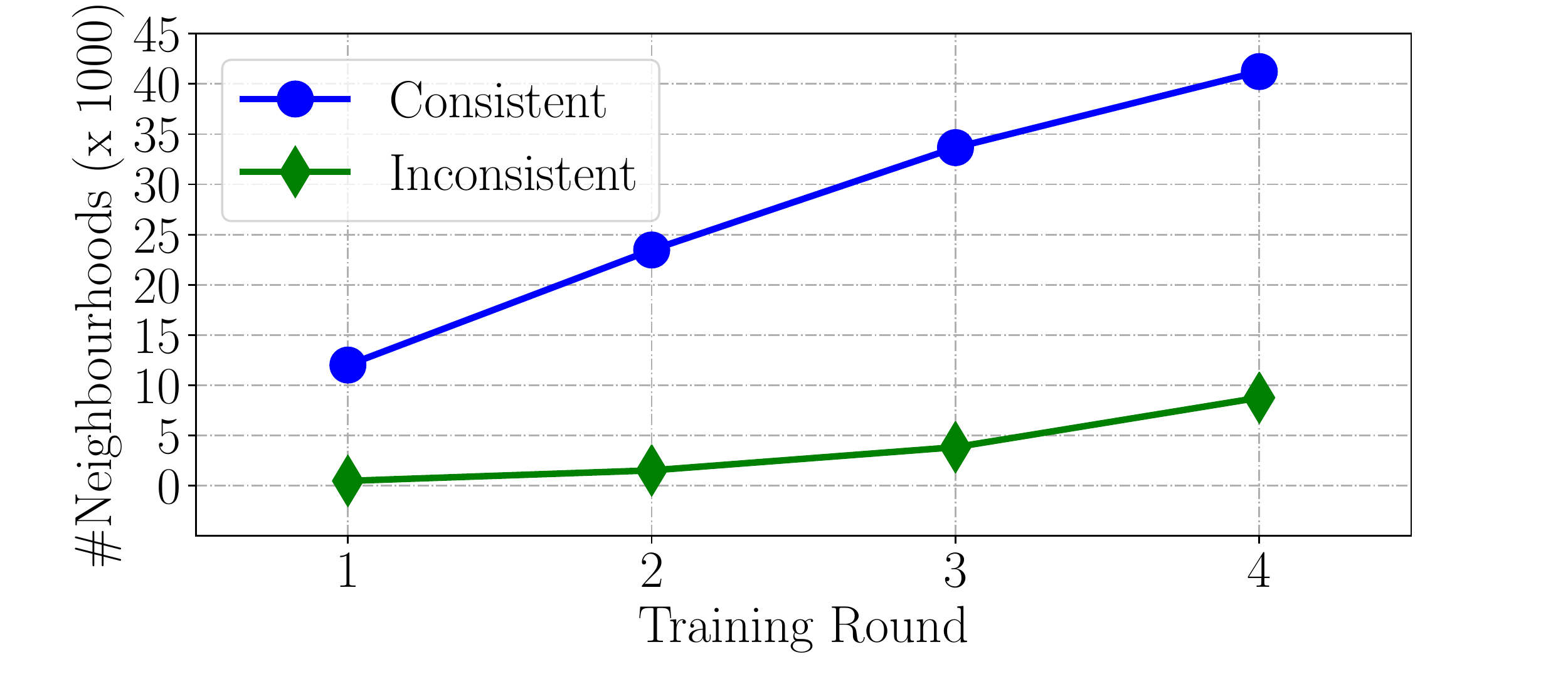}
    \end{center}
    \vskip -0.5cm
    \caption{Neighbourhood quality over rounds on CIFAR10.}
    \vspace{-0.3cm}
    \label{fig:neighbour_quality}
\end{figure}

{\bf Neighbourhood quality.}
We examined the class consistency of anchor neighbourhood
discovered throughout the curriculum rounds.
Figure \ref{fig:neighbour_quality} shows that
the numbers of both class consistent and inconsistent anchor neighbourhoods
increase along with the training rounds, and more importantly
the consistent ones raise much more rapidly.
This explains the performance advantages of
the AND model and the benefit of exploring progressive curriculum learning.

\begin{figure}[h]
\begin{center}
\includegraphics[width=0.95\linewidth]{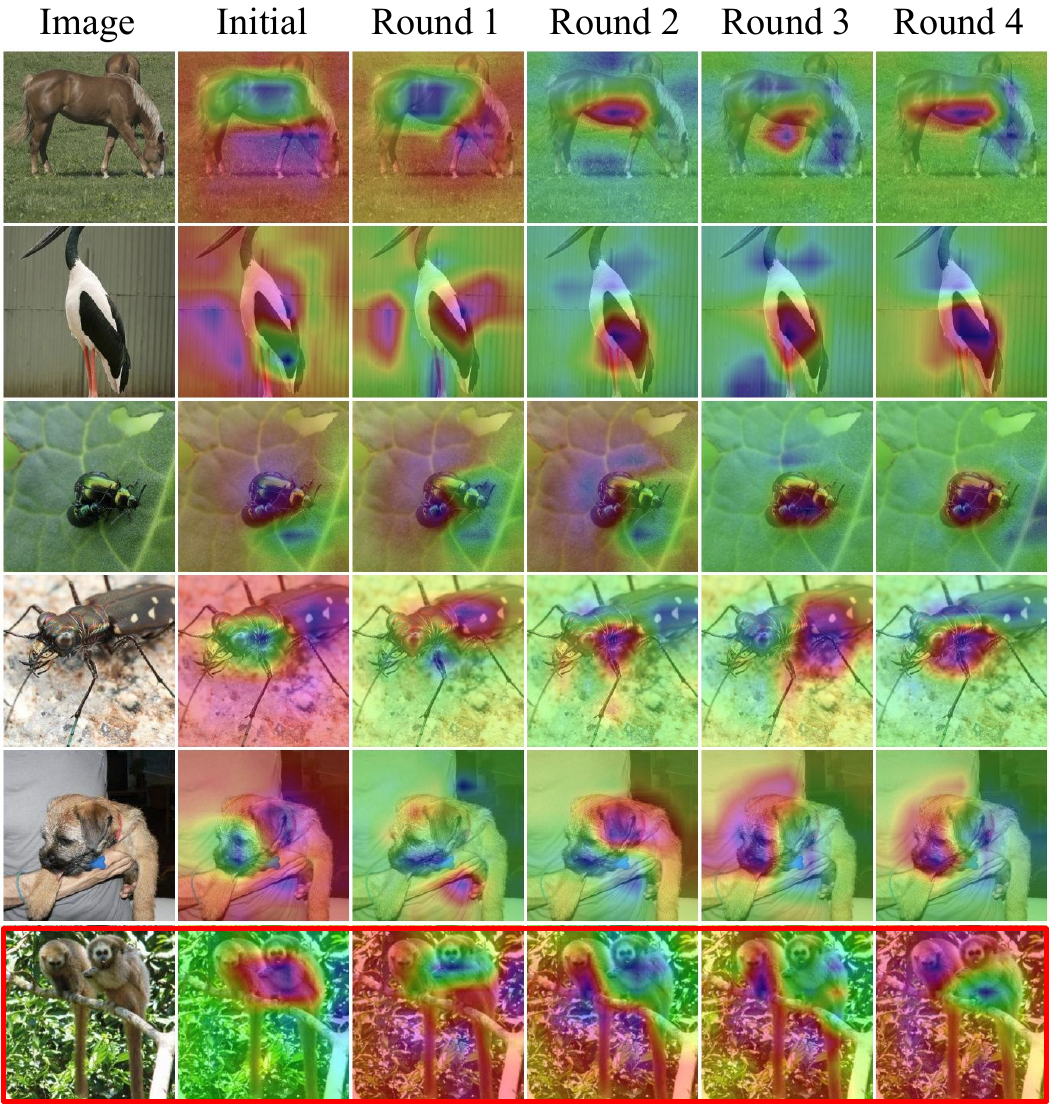}
\end{center}
    \vskip -0.5cm
   \caption{
    The evolving dynamics of model learning attention throughout the training rounds
    on six ImageNet classes.
    Red bounding box indicates a failure case.
   }
    \vspace{-0.5cm}
\label{fig:attention}
\end{figure}

{\bf Learning attention dynamics.}
To further understand how the AND benefits the feature representation learning,
we tracked the modelling attention by Grad-Cam~\cite{iccv2017selvaraju_grad-cam} to
visualise which parts of training images the model is focusing on 
throughout the curriculum rounds.
We have the following observations from Fig \ref{fig:attention}:
{\bf (1)} Often the model initially looks at class irrelevant image regions.
This suggests that sample specificity is a less effective
supervision signal for guiding model discriminative training.
{\bf (2)} In cases, the AND model is able to gradually 
shift the learning attention towards the class relevant parts
therefore yielding a more discriminative model.
{\bf (3)} The AND may fail to capture the object attention,
e.g. due to cluttered background and poor lighting condition.

\section{Conclusion}

In this work, we presented a novel Anchor Neighbourhood Discovery (AND) approach for unsupervised learning of discriminative deep network models
through class consistent neighbourhood discovery and supervision
in a progressive manner. 
With the AND model, we avoid the notorious grouping noises
whilst still preserving the intrinsic merits of clustering
for effective inference of the latent class decision boundaries.
Our method is also superior to the existing sample specificity learning 
strategy, due to the unique capability of propagating the self-discovered sample-to-sample class relationship information in end-to-end model optimisation. Extensive experiments on four image classification benchmarks show the modelling and performance superiority 
of the proposed AND method over a wide range of state-of-the-art unsupervised deep learning
models. We also provided in-depth component analysis to give insights
on the model 
advantages of the AND formulation.


\section*{Acknowledgements}
This work was partly supported by the China Scholarship Council,
Vision Semantics Limited,
the Royal Society Newton Advanced Fellowship Programme (NA150459),
and Innovate UK Industrial Challenge Project on Developing and Commercialising Intelligent
Video Analytics Solutions for Public Safety (98111-571149).


\bibliography{references}
\bibliographystyle{icml2019}

\end{document}